\documentclass[11pt,a4paper]{article}
\usepackage[hyperref]{emnlp2020}
\usepackage{times}
\usepackage{latexsym}

\usepackage{graphicx}
\usepackage{amsmath}
\usepackage{amsthm}
\usepackage{booktabs}
\usepackage{algorithm}
\usepackage{array}
\usepackage{subfigure}
\usepackage{multirow}
\usepackage{threeparttable}
\usepackage{tabularx}
\usepackage{booktabs}
\usepackage{colortbl}
\usepackage{xcolor}
\usepackage{stfloats}
\usepackage[noend]{algpseudocode}
\usepackage{mathtools}

\newenvironment{itemize*}%
 {\begin{itemize}%
  \setlength{\itemsep}{0pt}%
  \setlength{\parskip}{0pt}}%
 {\end{itemize}}
 
\usepackage{microtype}

\aclfinalcopy % Uncomment this line for the final submission
\title{BERT-ATTACK: Adversarial Attack Against BERT Using BERT}

 \author{
 Linyang Li, Ruotian Ma, Qipeng Guo, Xiangyang Xue, Xipeng Qiu\thanks{\ \  Corresponding author.}  \\
  Shanghai Key Laboratory of Intelligent Information Processing, Fudan University \\
  School of Computer Science, Fudan University \\
  825 Zhangheng Road, Shanghai, China \\
  \texttt{\{linyangli19,rtma19,qpguo16,xyxue,xpqiu\}@fudan.edu.cn}
  }

\date{}

\begin{document}
\maketitle

\begin{abstract}

Adversarial attacks for discrete data (such as texts) have been proved significantly more challenging than continuous data (such as images) since it is difficult to generate adversarial samples with gradient-based methods. Current successful attack methods for texts usually adopt heuristic replacement strategies on the character or word level, which remains
challenging to find the optimal solution in the massive space of possible combinations of replacements while preserving semantic consistency and language fluency.
In this paper, we propose \textbf{BERT-Attack}, a high-quality and effective method to generate adversarial samples using pre-trained masked language models exemplified by BERT.
We turn BERT against its fine-tuned models and other deep neural models in downstream tasks so that we can successfully mislead the target models to predict incorrectly.
Our method outperforms state-of-the-art attack strategies in both success rate and perturb percentage, while the generated adversarial samples are fluent and semantically preserved. 
Also, the cost of calculation is low, thus possible for large-scale generations.
The code is available at \url{https://github.com/LinyangLee/BERT-Attack}.

\end{abstract}

\section{Introduction}

Despite the success of deep learning, recent works have found that these neural networks are vulnerable to adversarial samples, which are crafted with small perturbations to the original inputs \cite{goodfellow2014explaining,kurakin2016adversarial,chakraborty2018adversarial}. That is, these adversarial samples are imperceptible to human judges while they can mislead the neural networks to incorrect predictions.
Therefore, it is essential to explore these adversarial attack methods since the ultimate goal is to make sure the neural networks are highly reliable and robust. While in computer vision fields, both attack strategies and their defense countermeasures are well-explored \cite{chakraborty2018adversarial}, the adversarial attack for text is still challenging due to the discrete nature of languages.
Generating of adversarial samples for texts needs to possess such qualities: (1) imperceptible to human judges yet misleading to neural models; (2) fluent in grammar and semantically consistent with original inputs.

Previous methods craft adversarial samples mainly based on specific rules~\cite{li2018textbugger,gao2018black,yang2018greedy,Alzantot,ren2019generating,jin2019textfooler,zang2020word}. Therefore, these methods are difficult to guarantee the fluency and semantically preservation in the generated adversarial samples at the same time.
Plus, these manual craft methods are rather complicated. They use multiple linguistic constraints like NER tagging or POS tagging. 
Introducing contextualized language models to serve as an automatic perturbation generator could make these rules designing much easier.

The recent rise of pre-trained language models, such as BERT \cite{bert}, push the performances of NLP tasks to a new level. On the one hand, the powerful ability of a fine-tuned BERT on downstream tasks makes it more challenging to be adversarial attacked~\cite{jin2019textfooler}.
On the other hand, BERT is a pre-trained masked language model on extremely large-scale unsupervised data and has learned general-purpose language knowledge. Therefore, BERT has the potential to generate more fluent and semantic-consistent substitutions for an input text. Naturally, both the properties of BERT motivate us to explore the possibility of attacking a fine-tuned BERT with another BERT as the attacker.

In this paper, we propose an effective and high-quality adversarial sample generation method: \textbf{BERT-Attack}, using BERT as a language model to generate adversarial samples.
The core algorithm of BERT-Attack is straightforward and consists of two stages: finding the vulnerable words in one given input sequence for the target model; then applying BERT in a semantic-preserving way to generate substitutes for the vulnerable words.
With the capability of BERT, the perturbations are generated considering the context around. Therefore, the perturbations are fluent and reasonable.
We use the masked language model as a perturbation generator and find perturbations that maximize the risk of making wrong predictions \cite{goodfellow2014explaining}. Differently from previous attacking strategies that require traditional single-direction language models as a constraint, we only need to inference the language model once as a perturbation generator rather than repeatedly using language models to score the generated adversarial samples in a trial and error process.

Experimental results show that the proposed BERT-Attack method successfully fooled its fine-tuned downstream model with the highest attack success rate compared with previous methods.
Meanwhile, the perturb percentage and the query number are considerably lower, while the semantic preservation is high.

To summarize our main contributions:
\begin{itemize}
    \item We propose a simple and effective method, named \textbf{BERT-Attack}, to effectively generate fluent and semantically-preserved adversarial samples that can successfully mislead state-of-the-art models in NLP, such as fine-tuned BERT for various downstream tasks.
    
    \item BERT-Attack has a higher attacking success rate and a lower perturb percentage with fewer access numbers to the target model compared with previous attacking algorithms, while does not require extra scoring models therefore extremely effective.
    
    % \item We can generate adversarial samples with BERT-Attack as a parallel dataset for further research on the robustness of NLP models.
\end{itemize}

\section{Related Work}

To explore the robustness of neural networks, adversarial attacks have been extensively studied for continuous data (such as images)~\cite{goodfellow2014explaining,nguyen2015deep,chakraborty2018adversarial}.
The key idea is to find a minimal perturbation that maximizes the risk of making wrong predictions.
This minimax problem can be easily achieved by applying gradient descent over the continuous space of images \cite{Miyato2017VirtualAT}.
However, adversarial attack for discrete data such as text remains challenging.

\textbf{Adversarial Attack for Text}

Current successful attacks for text usually adopt heuristic rules to modify the characters of a word
\cite{jin2019textfooler}, and substituting words with synonyms \cite{ren2019generating}.
\citet{li2018textbugger,gao2018black} apply perturbations based on word embeddings such as Glove \cite{pennington2014glove}, which is not strictly semantically and grammatically coordinated.
\citet{Alzantot} adopts language models to score the perturbations generated by searching for close meaning words in the word embedding space \cite{mrkvsic2016counter}, using a trial and error process to find possible perturbations, yet the perturbations generated are still not context-aware and heavily rely on cosine similarity measurement of word embeddings.
Glove embeddings do not guarantee similar vector space with cosine similarity distance, therefore the perturbations are less semantically consistent.
\citet{jin2019textfooler} apply a semantically enhanced embedding \cite{mrkvsic2016counter}, which is context unaware, thus less consistent with the unperturbed inputs.
\citet{liang2017deep} use phrase-level insertion and deletion, which produces unnatural sentences inconsistent with the original inputs, lacking fluency control.
To preserve semantic information, \citet{glockner2018breaking} replace words manually to break the language inference system \cite{bowman2015large}.
\citet{jia2017adversarial} propose manual craft methods to attack machine reading comprehension systems.
\citet{lei2019discrete} introduce replacement strategies using embedding transition.

Although the above approaches have achieved good results, there is still much room for improvement regarding the perturbed percentage, attacking success rate, grammatical correctness and semantic consistency, etc.
Moreover, the substitution strategies of these approaches are usually non-trivial, resulting in that they are limited to specific tasks.

\textbf{Adversarial Attack against BERT}

Pre-trained language models have become mainstream for many NLP tasks. Works such as \cite{Wallace2019Triggers,jin2019textfooler,pruthi2019combating} have explored these pre-trained language models from many different angles.
\citet{Wallace2019Triggers} explored the possible ethical problems of learned knowledge in pre-trained models.

% From our perspective, we take the idea of turning such language models against themselves. Therefore, we introduce a novel BERT-Attack algorithm to attack the fine-tuned models. 

\section{BERT-Attack}

Motivated by the interesting idea of turning BERT against BERT, we propose \textbf{BERT-Attack}, using the original BERT model to craft adversarial samples to fool the fine-tuned BERT model.

Our method consists of two steps: (1) finding the vulnerable words for the target model and then (2) replacing
them with the semantically similar and grammatically
correct words until a successful attack.

The most-vulnerable words are the keywords that help the target model make judgments. Perturbations over these words can be most beneficial in crafting adversarial samples. 
After finding which words that we are aimed to replace, we use masked language models to generate perturbations based on the top-K predictions from the masked language model.

\subsection{Finding Vulnerable Words}

Under the black-box scenario, the logit output by the target model (fine-tuned BERT or other neural models) is the only supervision we can get.
We first select the words in the sequence which have a high significance influence on the final output logit.

Let $S = [w_0, \cdots, w_i \cdots ]$ denote the input sentence, and  $o_y(S)$ denote the logit output by the target model for correct label $y$, the importance score $I_{w_i}$ is defined as
\begin{align}
    I_{w_i} = o_y(S) - o_y(S_{\backslash w_i}) , \label{eq:importance}
\end{align}
where $S_{\backslash w_i} = [w_0, \cdots, w_{i-1}, [\texttt{MASK}], w_{i+1}, \cdots]$ is the sentence after replacing $w_i$ with $[\texttt{MASK}]$.

Then we rank all the words according to the ranking score $I_{w_i}$ in descending order to create word list $L$.
We only take $\epsilon$ percent of the most important words since we tend to keep perturbations minimum.

This process maximizes the risk of making wrong predictions, which is previously done by calculating gradients in image domains. The problem is then formulated as replacing these most vulnerable words with semantically consistent perturbations.

\begin{algorithm*}[h]
\caption{BERT-Attack}\label{alg:bertattack}
\begin{algorithmic}[1]
\Procedure{Word Importance Ranking}{}
\State $ S = [w_0, w_1, \cdots]$ \textcolor[rgb]{0.00,0.50,0.00}{// input: tokenized sentence}
\State $Y \gets $ gold-label

\For{$ w_i$ in $S$}
\State calculate importance score $I_{w_i}$ using Eq. \ref{eq:importance}
\EndFor

\State select word list $L=[w_{top-1}, w_{top-2}, \cdots]$

\State \textcolor[rgb]{0.00,0.50,0.00}{// sort $S$ using $I_{w_i}$ in descending order and collect ${top}-K$ words}
\EndProcedure

\Procedure{Replacement using BERT}{}
\State $ H = [h_0, \cdots, h_n]$ \textcolor[rgb]{0.00,0.50,0.00}{// sub-word tokenized sequence of $S$}
%\State $V $ \textcolor[rgb]{0.00,0.50,0.00}{// vocabulary size of $\mathcal{M}$}
%\State $ P^{\in n*V} = \mathcal{M}(H)$
\State generate top-K candidates for all sub-words using BERT and get $P^{\in n\times K}$

\For{$w_j$ in $L$}

\If {$w_j$ is a whole word}
\State get candidate $C=Filter(P^{j})$
\State replace word $w_j$
\Else

\State get candidate $C$ using PPL ranking and Filter
\State replace sub-words $[h_j, \cdots, h_{j+t}]$
\EndIf

% \State Potential Sample Set $\gets \{\}$
\State Find Possible Adversarial Sample
\For {$c_k$ in C}
\State $S^{'} = [w_0, \cdots, w_{j-1}, c_k, \cdots]$ \textcolor[rgb]{0.00,0.50,0.00}{// attempt}
\If{$ \arg\!\max(o_{y}(S^{'})) != Y$ }
\State \textbf{\textit{return}} $S^{adv} = S^{'}$ \textcolor[rgb]{0.00,0.50,0.00}{// success attack}
\Else
% \State save $S^{'}$, logit $o_{S^{'}}$ 

\If{$o_{y}{(S^{'})} < o_{y}{(S^{adv})}$}
\State $S^{adv} = [w_0, \cdots, w_{j-1}, c, \cdots]$ \textcolor[rgb]{0.00,0.50,0.00}{// do one perturbation} 
\EndIf
\EndIf
\EndFor
% \State $\min(o_{S^{'}}) \text{in Potential Set}$ $ \gets$ Change $w_j$
% \State $c \gets \arg \min (o_{S^{'}})$ \textcolor[rgb]{0.00,0.50,0.00}{// select most successful perturbation $c$}
% \State $S^{adv} = [w_0, \cdots, c, \cdots]$ \textcolor[rgb]{0.00,0.50,0.00}{// do one perturbation}
\EndFor

\State \textbf{\textit{return} None}

\EndProcedure
\end{algorithmic}
\end{algorithm*}

\begin{figure}[t]
\centering
\includegraphics[width=0.95\linewidth]{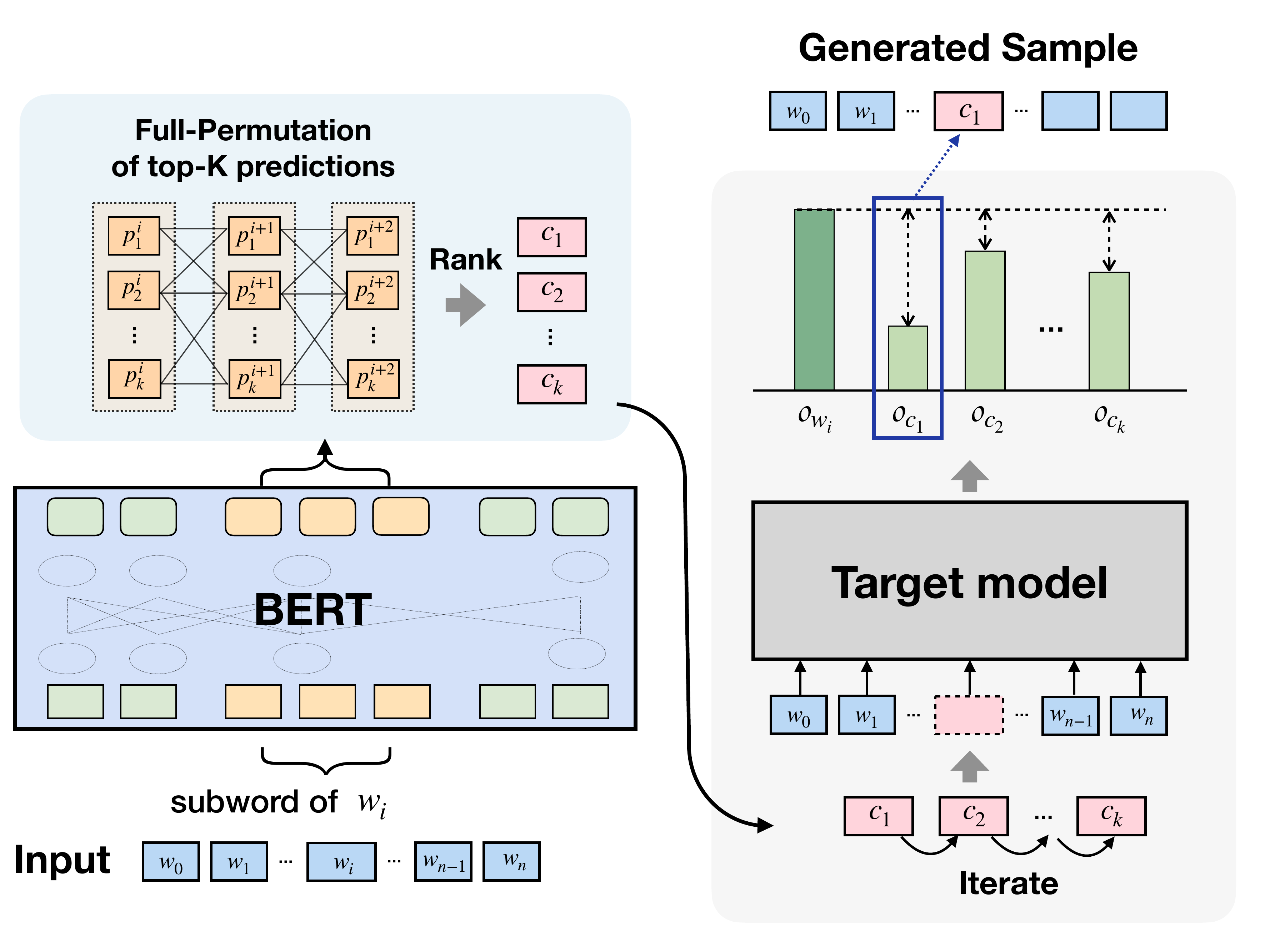}
\centering
\caption{One step of our replacement strategy.}
\label{fig:overall-structure}
\end{figure}

\subsection{Word Replacement via BERT}

After finding the vulnerable words, we iteratively replace the words in list $L$ one by one to find perturbations that can mislead the target model. Previous approaches usually use multiple human-crafted rules to ensure the generated example is semantically consistent with the original one and grammatically correct, such as a synonym dictionary \cite{ren2019generating}, POS checker \cite{jin2019textfooler}, semantic similarity checker \cite{jin2019textfooler}, etc.
\citet{Alzantot} applies a traditional language model to score the perturbed sentence at every attempt of replacing a word.

These strategies of generating substitutes are unaware of the context between the substitution positions (usually using language models to test the substitutions), thus are insufficient in fluency control and semantic consistency.
More importantly, using language models or POS checkers in scoring the perturbed samples is costly since this trial and error process requires massive inference time.

To overcome the lack of fluency control and semantic preservation by using synonyms or similar words in the embedding space, we leverage BERT for word replacement.
The genuine nature of the masked language model makes sure that the generated sentences are relatively fluent and grammar-correct, also preserve most semantic information, which is later confirmed by human evaluators.
Further, compared with previous approaches using rule-based perturbation strategies, the masked language model prediction is context-aware, thus dynamically searches for perturbations rather than simple synonyms replacing.

Different from previous methods using complicated strategies to score and constrain the perturbations, the contextualized perturbation generator generates minimal perturbations with only one forward pass. 
Without running additional neural models to score the sentence, the time-consuming part is accessing the target model only. 
Therefore the process is extremely efficient.

Thus, using the masked language model as a contextualized perturbation generator can be one possible solution to craft high-quality adversarial samples efficiently.

\subsubsection{Word Replacement Strategy}

As seen in Figure \ref{fig:overall-structure}, given a chosen word $w$ to be replaced, we apply BERT to predict the possible words that are similar to $w$ yet can mislead the target model.
Instead of following the masked language model settings, we do not mask the chosen word $w$ and use the original sequence as input, which can generate more semantic-consistent substitutes \cite{zhou-etal-2019-bert}.
For instance, given a sequence \textit{"I like the cat."}, if we mask the word \textit{cat}, it would be very hard for a masked language model to predict the original word \textit{cat} since it could be just as fluent if the sequence is \textit{"I like the dog."}.
Further, if we mask out the given word $w$, for each iteration we would have to rerun the masked language model prediction process which is costly.

Since BERT uses Bytes-Pair-Encoding (BPE) to tokenize the sequence $S= [w_0, \cdots, w_i, \cdots ]$ into sub-word tokens: $H = [h_0, h_1, h_2, \cdots]$, we need to align the chosen word to its corresponding sub-words in BERT.

Let $\mathcal{M}$ denote the BERT model, we feed the tokenized sequence $H$ into the BERT $\mathcal{M}$ to get output prediction $P = \mathcal{M}(H)$.
Instead of using the argmax prediction, we take the most possible $K$ predictions at each position, where $K$ is a hyper-parameter.

We iterate words that are sorted by word importance ranking process to find perturbations.
The BERT model uses BPE encoding to construct vocabularies. While most words are still single words, rare words are tokenized into sub-words.
Therefore, we treat single words and sub-words separately to generate the substitutes.

\paragraph{Single words}
For a single word $w_j$, we make attempts using the corresponding top-K prediction candidates ${P^j}$.
We first filter out stop words collected from NLTK; for sentiment classification tasks we filter out antonyms using synonym dictionaries \cite{mrkvsic2016counter} since BERT masked language model does not distinguish synonyms and antonyms. 
Then for given candidate $c_k$ we construct a perturbed sequence $H^{'} = [h_0, \cdots, h_{j-1}, c_k, h_{j+1} \cdots]$.
If the target model is already fooled to predict incorrectly, we break the loop to obtain the final adversarial sample $H^{adv}$; otherwise, we select from the filtered candidates to pick one best perturbation and turn to the next word in word list $L$.

\paragraph{Sub-words}
For a word that is tokenized into sub-words in BERT, we cannot obtain its substitutes directly.
Thus we use the perplexity of sub-word combinations to find suitable word substitutes from predictions in the sub-word level.
Given sub-words $[h_0, h_1, \cdots, h_t]$ of word $w$, we list all possible combinations from the prediction $P^{\in t\times K}$ from $\mathcal{M}$, which is $K^t$ sub-word combinations, we can convert them back to normal words by reversing the BERT tokenization process.
We feed these combinations into the BERT-MLM to get the perplexity of these combinations.
Then we rank the perplexity of all combinations to get the top-K combinations to find the suitable sub-word combinations.

Given the suitable perturbations, we replace the original word with the most likely perturbation and repeat this process by iterating the importance word ranking list to find the final adversarial sample.
In this way, we acquire the adversarial samples $S^{adv}$ effectively since we only iterate the masked language model once and do perturbations using the masked language model without other checking strategies.

We summarize the two-step BERT-Attack process in Algorithm \ref{alg:bertattack}.

\section{Experiments}

\subsection{Datasets}

We apply our method to attack different types of NLP tasks in the form of text classification and natural language inference.
Following \citet{jin2019textfooler}, we evaluate our method on 1k test samples randomly selected from the test set of the given task which are the same splits used by \citet{Alzantot,jin2019textfooler}.
The GA method only uses a subset of 50 samples in the FAKE, IMDB dataset.
% We implement Textfooler\footnote{https://github.com/jind11/TextFooler} and GA \footnote{https://github.com/jind11/TextFooler} as comparative methods.

\paragraph{Text Classification}

We use different types of text classification tasks to study the effectiveness of our method.

\begin{itemize*}
    \item \textbf{Yelp} Review classification dataset, containing. Following \citet{zhang2015character}, we process the dataset to construct a polarity classification task.

    \item \textbf{IMDB} Document-level movie review dataset, where the average sequence length is longer than the Yelp dataset. We process the dataset into a polarity classification task \footnote{ https://datasets.imdbws.com/}.

    \item \textbf{AG's News} Sentence level news-type classification dataset, containing 4 types of news: World, Sports, Business, and Science.

    \item \textbf{FAKE} Fake News Classification dataset, detecting whether a news document is fake from Kaggle Fake News Challenge \footnote{ https://www.kaggle.com/c/fake-news/data}.

\end{itemize*}

\paragraph{Natural Language Inference}

\begin{itemize*}

    \item \textbf{SNLI} Stanford language inference task \cite{bowman2015large}. Given one premise and one hypothesis, and the goal is to predict if the hypothesis is entailment, neural, or contradiction of the premise.

    \item \textbf{MNLI} Language inference dataset on multi-genre texts, covering transcribed speech, popular fiction, and government reports \cite{mnli}, which is more complicated with diversified written and spoken style texts, compared with the SNLI dataset, including eval data matched with training domains and eval data mismatched with training domains.

\end{itemize*}

\subsection{Automatic Evaluation Metrics}

To measure the quality of the generated samples, we set up various automatic evaluation metrics.
The success rate, which is the counter-part of after-attack accuracy, is the core metric measuring the success of the attacking method.
Meanwhile, the perturbed percentage is also crucial since, generally, less perturbation results in more semantic consistency.
Further, under the black-box setting, queries of the target model are the only accessible information. Constant queries for one sample is less applicable. Thus query number per sample is also a key metric.
As used in TextFooler \cite{jin2019textfooler}, we also use Universal Sentence Encoder \cite{cer2018universal} to measure the semantic consistency between the adversarial sample and the original sequence.
To balance between semantic preservation and attack success rate, we set up a threshold of semantic similarity score to filter the less similar examples.

\begin{table*}[ht]\setlength{\tabcolsep}{2pt}
    \centering \small
    \begin{tabular}{cccccccc}
        \toprule
        \textbf{Dataset}& \textbf{Method} & \textbf{Original Acc} & \textbf{Attacked Acc} & \textbf{Perturb \%} & \textbf{Query Number} & \textbf{Avg Len}& \textbf{Semantic Sim} \\
        \midrule
        \multirow{3}*{\bfseries Fake} &
        BERT-Attack(ours)& \multirow{3}*{97.8} & \textbf{15.5} & \textbf{1.1} & \textbf{1558} & \multirow{3}*{885} & \textbf{0.81}\\
        \cmidrule{2-2} &TextFooler\cite{jin2019textfooler}& & 19.3 & 11.7 & 4403 &  & 0.76\\
        \cmidrule{2-2}
        &GA\cite{Alzantot}& & 58.3 & 1.1 & 28508 &  & -\\
        \midrule
        \multirow{3}*{\bfseries Yelp} &
        BERT-Attack(ours)& \multirow{3}*{95.6} & \textbf{5.1} & \textbf{4.1} & \textbf{273} & \multirow{3}*{157} & \textbf{0.77}\\
        \cmidrule{2-2} &TextFooler&  & \text{6.6} & \text{12.8} & \text{743} & \text{} & \text{0.74}\\
         \cmidrule{2-2} &GA& &  31.0 & 10.1 & 6137  &  & -\\
        \midrule
        \multirow{3}*{\bfseries IMDB} &
        BERT-Attack(ours)& \multirow{3}*{90.9} & \textbf{11.4} & \textbf{4.4} & \textbf{454} & \multirow{3}*{215} & \textbf{0.86}\\
        \cmidrule{2-2} &TextFooler&  & \text{13.6} & \text{6.1} & \text{1134} & \text{} & \textbf{0.86}\\
        \cmidrule{2-2} &GA &  & 45.7 & 4.9 & 6493 & \text{}& -\\
        \midrule
        \multirow{3}*{\bfseries AG} &
        BERT-Attack(ours)& \multirow{3}*{94.2} & \textbf{10.6} & \textbf{15.4} & \textbf{213} & \multirow{3}*{43} & \textbf{0.63}\\
        \cmidrule{2-2} &TextFooler&  & \text{12.5} & \text{22.0} & \text{357} & \text{} & \text{0.57}\\
        \cmidrule{2-2} &GA&  & \text{51} & \text{16.9} & \text{3495} & \text{} & -\\
        \midrule

       \multirow{3}*{\bfseries SNLI} &
        BERT-Attack(ours)& \multirow{3}*{89.4(H/P)} & \text{7.4/}\textbf{16.1} & \textbf{12.4/9.3} & \textbf{16/30} & \multirow{3}*{8/18} & \text{0.40/}\textbf{0.55}\\
        \cmidrule{2-2} &TextFooler& & \text{\textbf{4.0}/20.8} & \text{18.5/33.4} & \text{60/142} & \text{} & \text{\textbf{0.45}/0.54}\\
        \cmidrule{2-2} &GA& & \text{14.7/-} & \text{20.8/-} & \text{613/-} & \text{} & -\\
        \midrule
       \multirow{3}*{\bfseries MNLI} &
        BERT-Attack(ours)& \multirow{3}*{85.1(H/P)} & \textbf{7.9/11.9} & \textbf{8.8/7.9} & \textbf{19/44} & \multirow{3}*{11/21} & \text{0.55/\textbf{0.68}}\\
        \cmidrule{2-2} matched&TextFooler& & \text{9.6/25.3} & \text{15.2/26.5} & \text{78/152} & \text{} & \text{\textbf{0.57}/0.65}\\
        \cmidrule{2-2} &GA& & \text{21.8/-} & \text{18.2/-} & \text{692/-} & \text{} & \text{-}\\
        \midrule
       \multirow{3}*{\bfseries MNLI} &
        BERT-Attack(ours)& \multirow{3}*{82.1(H/P)} & \textbf{7/13.7} & \textbf{8.0/7.1} & \textbf{24/43} & \multirow{3}*{12/22} & \text{0.53/}\textbf{0.69}\\
        \cmidrule{2-2} mismatched&TextFooler& & \text{8.3/22.9} & \text{14.6/24.7} & \text{86/162} & \text{} & \text{\textbf{0.58}/0.65}\\
        \cmidrule{2-2} &GA& & \text{20.9/-} & \text{19.0/-} & \text{737/-} & \text{} & -\\
        \midrule

    \end{tabular}
    \caption{Results of attacking against various fine-tuned BERT models. TextFooler is the state-of-the-art baseline. For MNLI task, we attack the hypothesis(H) or premises(P) separately.
    }

    \label{tab:attack-results}
\end{table*}

\subsection{Attacking Results}

As shown in Table \ref{tab:attack-results}, the BERT-Attack method successfully fool its downstream fine-tuned model.
In both text classification and natural language inference tasks, the fine-tuned BERTs fail to classify the generated adversarial samples correctly.

The average after-attack accuracy is lower than 10\%, indicating that most samples are successfully perturbed to fool the state-of-the-art classification models.
Meanwhile, the perturb percentage is less than 10 \%, which is significantly less than previous works.

Further, \textbf{BERT-Attack} successfully attacked all tasks listed, which are in diversified domains such as News classification, review classification, language inference in different domains.
The results indicate that the attacking method is robust in different tasks.
Compared with the strong baseline introduced by \citet{jin2019textfooler}\footnote{https://github.com/jind11/TextFooler} and \citet{Alzantot}\footnote{https://github.com/QData/TextAttack}, the BERT-Attack method is more efficient and more imperceptible. The query number and the perturbation percentage of our method are much less.

We can observe that it is generally easier to attack the review classification task since the perturb percentage is incredibly low.
BERT-Attack can mislead the target model by replacing a handful of words only.
Since the average sequence length is relatively long,  the target model tends to make judgments by only a few words in a sequence, which is not the natural way of human prediction.
Thus, the perturbation of these keywords would result in incorrect prediction from the target model, revealing the vulnerability of it.

\subsection{Human Evaluations}

For further evaluation of the generated adversarial samples, we set up human evaluations to measure the quality of the generated samples in fluency and grammar as well as semantic preservation.

We ask human judges to score the grammar correctness of the mixed sentences of generated adversarial samples and original sequences, scoring from 1-5 following \citet{jin2019textfooler}.
Then we ask human judges to make predictions in a shuffled mix of original and adversarial texts.
We use the IMDB dataset and the MNLI dataset, and for each task, we select 100 samples of both original and adversarial samples for human judges. 
We ask three human annotators to evaluate the examples. For label prediction, we take the majority class as the predicted label, and for semantic and grammar check we use an average score among the annotators. 

Seen in Table \ref{tab:humaneval}, the semantic score and the grammar score of the adversarial samples are close to the original ones. 
MNLI task is a sentence pair prediction task constructed by human crafted hypotheses based on the premises, therefore original pairs share a considerable amount of same words.
Perturbations on these words would make it difficult for human judges to predict correctly therefore the accuracy is lower than simple sentence classification tasks.

\begin{table}[ht]\setlength{\tabcolsep}{2pt}\small
    \centering
    \begin{tabular}{ccccccc}
        \toprule
        \multicolumn{3}{c}{\textbf{Dataset}}   & \textbf{Accuracy } & \textbf{Semantic } & \textbf{Grammar }\\
        \midrule
        \multirow{2}*{\textbf{MNLI}} & &Original  &0.90 & 3.9 & 4.0 \\
        & & Adversarial &0.70 & 3.7 & 3.6 \\
        \midrule
        \multirow{2}*{\textbf{IMDB}} && Original & 0.91 & 4.1 & 3.9\\
        & & Adversarial & 0.85 & 3.9 & 3.7 \\
        \bottomrule

    \end{tabular}
    \caption{Human-Evaluation Results.  
    }

    \label{tab:humaneval}
\end{table}

\subsection{BERT-Attack against Other Models}

The BERT-Attack method is also applicable in attacking other target models, not limited to its fine-tuned model only.
As seen in Table \ref{tab:other-models}, the attack is successful against LSTM-based models, indicating that BERT-Attack is feasible for a wide range of models.
Under BERT-Attack, the ESIM model is more robust in the MNLI dataset.
We assume that encoding two sentences separately gets higher robustness.
In attacking BERT-large models, the performance is also excellent, indicating that BERT-Attack is successful in attacking different pre-trained models not only against its own fine-tuned downstream models.

\begin{table}[ht]\setlength{\tabcolsep}{2pt}\small
    \centering
    \begin{tabular}{ccccc}
        \toprule
        \textbf{Dataset}& \textbf{Model} & \textbf{Ori Acc} & \textbf{Atk Acc} & \textbf{Perturb \%} \\
        \midrule
        \multirow{2}*{\bfseries IMDB} &
        Word-LSTM & 89.8 & 10.2 & 2.7\\
        \cmidrule{2-2} &BERT-Large & 98.2 & 12.4 & 2.9 \\
        \midrule
        \multirow{2}*{\bfseries Yelp} &
        Word-LSTM &  {96.0} & 1.1 & 4.7 \\
        \cmidrule{2-2} &BERT-Large & 97.9 & 8.2 & 4.1\\
        \midrule
        \multirow{2}*{\bfseries MNLI} &
        ESIM  & 76.2 & 9.6 & 21.7\\
        \cmidrule{2-2} matched&BERT-Large & 86.4 & 13.2 & 7.4 \\
        \bottomrule
    \end{tabular}
    \caption{BERT-Attack against other models.
    }
    \label{tab:other-models}
\end{table}

\begin{figure}[ht]
\centering
\includegraphics[width=1.0\linewidth]{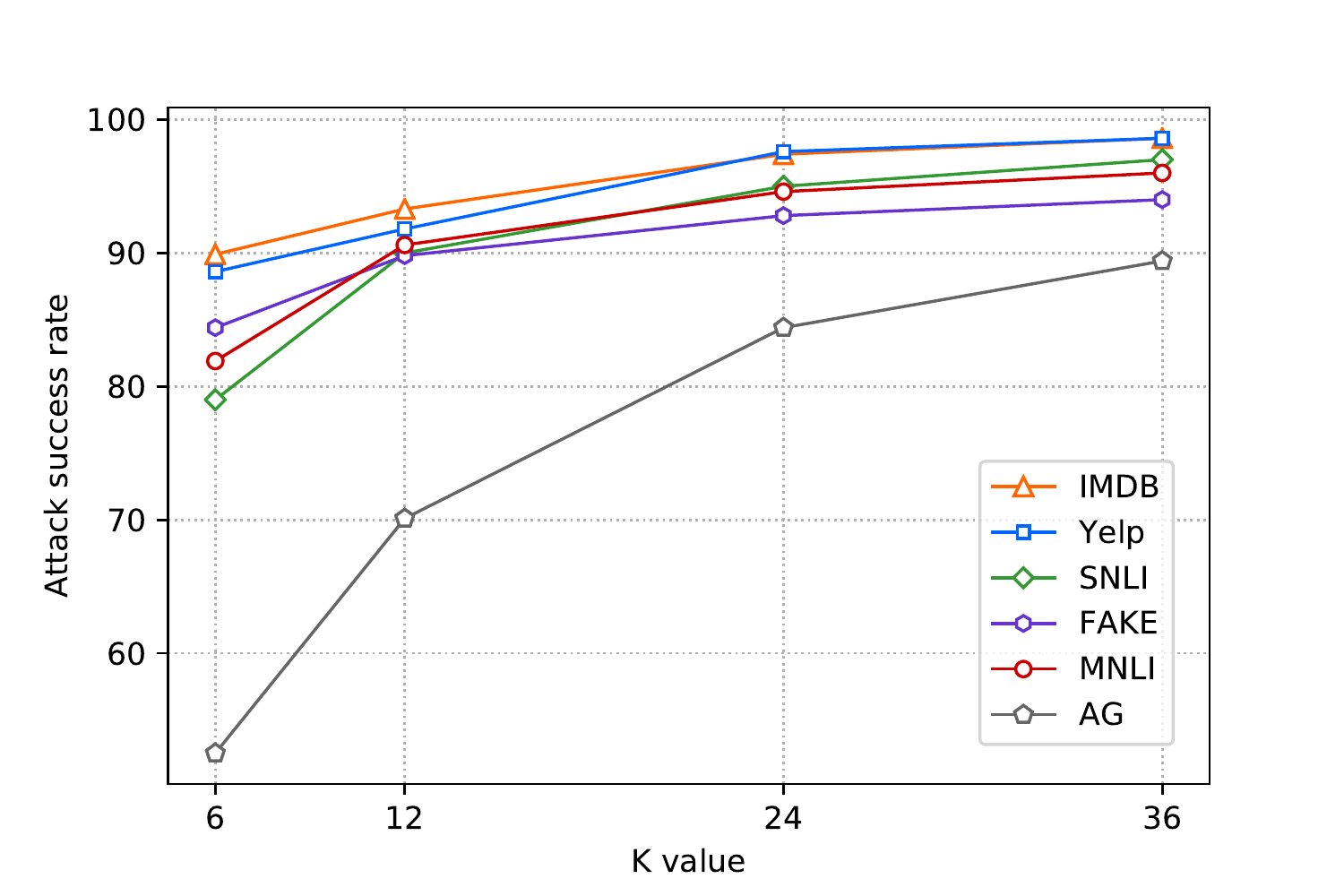}
\centering
\caption{Using different candidate number $K$ in the attacking process.}
\label{fig:k-num}
\end{figure}

\begin{table}[htb]\setlength{\tabcolsep}{2pt}\small
    \centering
    \begin{tabular}{cccccc}
        \toprule
        \textbf{Dataset}& \textbf{Method} & \textbf{Ori Acc} &\textbf{Atk Acc} & \textbf{Queries \%}  \\
        \midrule
        \multirow{2}*{\bfseries IMDB} &
        Fixed-$K$ &  90.9 & 11.4 & 454\\
        \cmidrule{2-2} & With Threshold & 90.9 & 12.4 & 440 \\
        \bottomrule

    \end{tabular}
    \caption{Flexible Candidates Using a threshold to cut off unsuitable candidates.
    }

    \label{tab:flexible-candidates}
\end{table}

% additional exp

\section{Ablations and Discussions}

\subsection{Importance of Candidate Numbers}

The candidate pool range is the major hyper-parameter used in the BERT-Attack algorithm.
As seen in Figure \ref{fig:k-num},  the attack rate is rising along with the candidate size increasing.
Intuitively, a larger $K$ would result in less semantic similarity.
However, the semantic measure via Universal Sentence Encoder is maintained in a stable range, (experiments show that semantic similarities drop less than 2\%),
indicating that the candidates are all reasonable and semantically consistent with the original sentence.

Further, a fixed candidate number could be rigid in practical usage, so we run a test using a threshold to cut off candidates that are less possible as a plausible perturbation.

As seen in Table \ref{tab:flexible-candidates}, when using a flexible threshold to cut off unsuitable candidates, the attacking process has a lower query number. 
This indicates that some candidates predicted by the masked language model with a lower prediction score may not be meaningful so skipping these candidates can save the unnecessary queries.

\subsection{Importance of Sequence Length}

The BERT-Attack method is based on the contextualized masked language model. Thus the sequence length plays an important role in the high-quality perturbation process.
As seen, instead of the previous methods focusing on attacking the hypothesis of the NLI task, we aim at premises whose average length is longer.
This is because we believe that contextual replacement would be less reasonable when dealing with extremely short sequences.
To avoid such a problem, we believe that many word-level synonym replacement strategies can be combined with BERT-Attack, allowing the BERT-Attack method to be more applicable.

\begin{table}[ht]\setlength{\tabcolsep}{2pt}\small
    \centering
    \begin{tabular}{cccccc}
        \toprule
        \textbf{Dataset}& \textbf{Method} & \textbf{Ori Acc} &\textbf{Atk Acc} & \textbf{Perturb \%}  \\
        \midrule
        \multirow{2}*{\bfseries MNLI} &
        BERT-Atk & 85.1 & 7.9 & 8.8 \\
        \cmidrule{2-2} matched&+Adv Train & 84.6 & 23.1& 10.5 \\
        \bottomrule

    \end{tabular}
    \caption{Adversarial training results.
    }

    \label{tab:adverse-training}
\end{table}

\begin{table}[ht]\setlength{\tabcolsep}{1pt}\small
    \centering
    \begin{tabular}{ccccc}
        \toprule
        \textbf{Dataset}& \textbf{Model} & \text{LSTM} & \text{BERT-base} & \text{BERT-large} \\
        \midrule
        \multirow{3}*{\bfseries IMDB} &
        Word-LSTM & - & 0.78 & 0.75 \\
        \cmidrule{2-2} &BERT-base & 0.83 & - & 0.71 \\
        \cmidrule{2-2} &BERT-large & 0.87 & 0.86 & - \\
        \toprule
        \textbf{Dataset}& \textbf{Model} & \text{ESIM} & \text{BERT-base} & \text{BERT-large} \\
        \midrule
        \multirow{3}*{\bfseries MNLI} &
        ESIM & - & 0.59 & 0.60 \\
        \cmidrule{2-2} &BERT-base & 0.60 & - & 0.45 \\
        \cmidrule{2-2} &BERT-large & 0.59 & 0.43 & -\\
        \bottomrule

    \end{tabular}
    \caption{Transferability analysis using attacked accuracy as the evaluation metric. The column is the target model used in attack, and the row is the tested model. 
    }

    \label{tab:transferability}
\end{table}

\subsection{Transferability and Adversarial Training}

To test the transferability of the generated adversarial samples, we take samples aimed at different target models to attack other target models.
Here, we use BERT-base as the masked language model for all different target models.
As seen in Table \ref{tab:transferability}, samples are transferable in NLI task while less transferable in text classification.

Meanwhile, we further fine-tune the target model using the generated adversarial samples from the train set and then test it on the test set used before.
As seen in Table \ref{tab:adverse-training},
generated samples used in fine-tuning help the target model become more robust while accuracy is close to the model trained with clean datasets.
The attack becomes more difficult, indicating that the model is harder to be attacked.
Therefore, the generated dataset can be used as additional data for further exploration of making neural models more robust.

\begin{table}[ht]\setlength{\tabcolsep}{2pt}\small
    \centering
    \begin{tabular}{ccccc}
        \toprule
        \textbf{Dataset}& \textbf{Model} & \textbf{Atk Acc} & \textbf{Perturb \%} & \textbf{Semantic} \\
        \midrule
        \multirow{2}*{\bfseries Yelp} &
        BERT-Atk & 5.1 & 4.1 & 0.77\\
        \cmidrule{2-2} &w/o sub-word & 7.1 & 4.3 & 0.74\\
        \midrule
        \multirow{2}*{\bfseries MNLI} &
        BERT-Atk & 11.9 & 7.9 & 0.68\\
        \cmidrule{2-2} &w/o sub-word & 14.7 & 9.3 & 0.63\\
        \bottomrule

    \end{tabular}
    \caption{Effects on sub-word level attack.}

    \label{tab:bpe}
\end{table}

\subsection{Effects on Sub-Word Level Attack}

BPE method is currently the most efficient way to deal with a large number of words, as used in BERT.
We establish a comparative experiment where we do not use the sub-word level attack. That is we skip those words that are tokenized with multiple sub-words.

As seen in Table \ref{tab:bpe}, using the sub-word level attack can achieve higher performances, not only in higher attacking success rate but also in less perturbation percentage.

\begin{table}[ht]\setlength{\tabcolsep}{2pt}\small
    \centering
    \begin{tabular}{ccccc}
        \toprule
        \textbf{Dataset}& \textbf{Method} & \textbf{Atk Acc} & \textbf{Perturb \%} & \textbf{Semantic} \\
\midrule
        \multirow{3}*{\bfseries MNLI} &
        MIR & 7.9 & 8.8 & 0.68 \\
        \cmidrule{2-2} \multirow{2}*{matched}&Random &
        20.2& 12.2& 0.60\\
        \cmidrule{2-2} &LIR & 27.2 & 15.0&0.60 \\
        \bottomrule
    \end{tabular}
    \caption{Most Importance Ranking (MIR) vs Least Importance Ranking (LIR)
    }

    \label{tab:word-importance}
\end{table}

\subsection{Effects on Word Importance Ranking }

Word importance ranking strategy is supposed to find keys that are essential to NN models, which is very much like calculating the maximum risk of wrong predictions in the FGSM algorithm \cite{goodfellow2014explaining}.
When not using word importance ranking, the attacking algorithm is less successful.

\begin{table}[ht]\setlength{\tabcolsep}{2pt}\small
    \centering 
    \begin{tabular}{ccccc}
        \toprule
        \textbf{Dataset}& \textbf{Method} & \textbf{Runtime(s/sample)}  \\
\midrule
        \multirow{3}*{\bfseries IMDB} &
        BERT-Attack(w/o BPE) & 14.2   \\
        \cmidrule{2-2}
        & BERT-Attack(w/ BPE) & 16.0   \\
        \cmidrule{2-2} 
        \multirow{2}*{}&Textfooler\cite{jin2019textfooler} &
        42.4\\
        \cmidrule{2-2} & GA\cite{Alzantot} & 2582.0  \\
        \bottomrule
    \end{tabular}
    \caption{Runtime comparison.
    }

    \label{tab:run-time}
\end{table}

\newcolumntype{L}[1]{>{\raggedright\let\newline\\\arraybackslash\hspace{0pt}}m{#1}}
\begin{table*}[ht]\setlength{\tabcolsep}{2pt}\small
    
    \begin{tabular}{lllL{2cm}llll}
        \toprule 
        \textbf{Dataset}&&&&&\text{Label} \\
        \midrule

        \multirow{3}{*}{\bfseries MNLI} &
        {Ori} &
        {\textcolor[rgb]{0.00,0.00,1.00}{Some} rooms have balconies .}
        &\multicolumn{1}{c}{Hypothesis}&\multicolumn{1}{l}{All of the rooms have balconies off of them .}
        &Contradiction\\
        \cmidrule{2-6} &\multirow{1}*{Adv} &
        {\textcolor{red}{Many} rooms have balconies .}
        &\multicolumn{1}{c}{Hypothesis}&\multicolumn{1}{l}{\text{All of the rooms have balconies off of them .}}&\textcolor{red}{Neutral}\\

        \midrule

        \multirow{6}{*}{\bfseries IMDB} &
        \multirow{3}*{Ori  } &
        \multicolumn{3}{l}{it is hard for a lover of the novel northanger abbey to sit through this bbc adaptation and to }
        &Negative\\
        &&\multicolumn{3}{l}{keep from throwing objects at the tv screen... why are so many facts concerning the tilney  }&& \\
        &&\multicolumn{3}{l}{family and mrs . tilney ' s death altered unnecessarily ? to make the \textcolor[rgb]{0.00,0.07,1.00}{story} more ` horrible ? ' }\\

        \cmidrule{2-6} &\multirow{3}*{Adv} &
        \multicolumn{3}{l}{it is hard for a lover of the novel northanger abbey to sit through this bbc adaptation and to}
        &\textcolor{red}{Positive}\\
        &&\multicolumn{3}{l}{keep from throwing objects at the tv screen... why are so many facts concerning the tilney }&& \\
        &&\multicolumn{3}{l}{family and mrs . tilney ' s death  altered unnecessarily ? to make the \textcolor{red}{plot} more ` horrible ? ' }\\
        \midrule

        \multirow{6}*{\bfseries IMDB} &
        \multirow{3}*{Ori} &
        \multicolumn{3}{l}{i first seen this movie in the early 80s .. it really had nice picture quality too . anyways , i 'm  }
        &Positive\\
        &&\multicolumn{3}{l}{glad i found this movie again ... the part i loved best was when he hijacked the car from this}&& \\
        && \multicolumn{3}{l}{poor guy... this is a movie i could watch over and over again . i {\textcolor[rgb]{0.00,0.00,1.00}{highly}} recommend it .}&& \\

        \cmidrule{2-6} &\multirow{3}*{Adv} &
        \multicolumn{3}{l}{i first seen this movie in the early 80s .. it really had nice picture quality too . anyways , i 'm  }
        &\textcolor{red}{Negative}\\
        &&\multicolumn{3}{l}{glad i found this movie again ... the part i loved best was when he hijacked the car from this}&& \\
        && \multicolumn{3}{l}{poor guy... this is a movie i could watch over and over again . i \textcolor{red}{inordinately} recommend it .}&& \\

        \midrule

    \end{tabular}
    \caption{Some generated adversarial samples. Origin label is the correct prediction while \textcolor{red}{label} is adverse prediction. Only red color parts are perturbed. We only attack premises in MNLI task. Text in FAKE dataset and IMDB dataset is cut to fit in the table. Original text contains more than 200 words.
    }

    \label{tab:samples}
\end{table*}

\subsection{Runtime Comparison}

Since BERT-Attack does not use language models or sentence encoders to measure the output sequence during the generation process, also, the query number is lower, therefore the runtime is faster than previous methods.
As seen in Table \ref{tab:run-time}, BERT-Attack is much faster than generic algorithm \cite{Alzantot} and 3 times faster then Textfooler.

\subsection{Examples of Generated Adversarial Sentences}

As seen in Table \ref{tab:samples}, the generated adversarial samples are semantically consistent with its original input, while the target model makes incorrect predictions.
In both review classification samples and language inference samples, the perturbations do not mislead human judges.

\section{Conclusion}

In this work, we propose a high-quality and effective method \textbf{BERT-Attack} to generate adversarial samples using BERT masked language model.
Experiment results show that the proposed method achieves a high success rate while maintaining a minimum perturbation.
Nevertheless, candidates generated from the masked language model can sometimes be antonyms or irrelevant to the original words, causing a semantic loss.
Thus, enhancing language models to generate more semantically related perturbations can be one possible solution to perfect BERT-Attack in the future.

\section{Acknowledgement}
We would like to thank the anonymous reviewers for their valuable comments.
We are thankful for the help of Demin Song, Hang Yan and Pengfei Liu.
This work was supported by the National Natural Science Foundation of China (No. 61751201, 62022027 and 61976056), Shanghai Municipal Science and Technology Major Project (No. 2018SHZDZX01) and ZJLab.

\bibliography{emnlp2020}
\bibliographystyle{acl_natbib}

% \appendix

% \section{Appendices}
% \label{sec:appendix}

\end{document}